\documentclass{article}
\usepackage{spconf,amsmath,graphicx}
\usepackage{multirow}
\usepackage{arydshln}
\usepackage{amsmath,amsfonts}
\usepackage{enumitem}
\usepackage{caption}
\usepackage{subfig}

\usepackage{xcolor}
\usepackage{hyperref}
\usepackage{cleveref}
\def\x{{\mathbf x}}
\def\e{{\mathbf e}}
\def\W{{\mathbf W}}
\def\v{{\mathbf v}}
\def\r{{\mathbf r}}
\def\z{{\mathbf z}}
\def\y{{\mathbf y}}

\title{Task Indicating Transformer for Task-conditional Dense Predictions}
\name{Yuxiang Lu, Shalayiding Sirejiding, Bayram Bayramli, Suizhi Huang, Yue Ding, Hongtao Lu$^*$ \thanks{This paper is supported by National Nature Science Foundation of China (62176155, 62066002), Shanghai Municipal Science and Technology Major Project (2021SHZDZX0102). Hongtao Lu is also with MOE Key Lab of Artificial Intelligence, AI Institute, Shanghai Jiao Tong University.}}
\address{Department of Computer Science and Engineering, Shanghai Jiao Tong University, Shanghai, China}

\begin{document}
\ninept
\maketitle
\begin{abstract}
The task-conditional model is a distinctive stream for efficient multi-task learning. Existing works encounter a critical limitation in learning task-agnostic and task-specific representations, primarily due to shortcomings in global context modeling arising from CNN-based architectures, as well as a deficiency in multi-scale feature interaction within the decoder.
In this paper, we introduce a novel task-conditional framework called Task Indicating Transformer (TIT) to tackle this challenge. Our approach designs a Mix Task Adapter module within the transformer block, which incorporates a Task Indicating Matrix through matrix decomposition, thereby enhancing long-range dependency modeling and parameter-efficient feature adaptation by capturing intra- and inter-task features. Moreover, we propose a Task Gate Decoder module that harnesses a Task Indicating Vector and gating mechanism to facilitate adaptive multi-scale feature refinement guided by task embeddings. Experiments on two public multi-task dense prediction benchmarks, NYUD-v2 and PASCAL-Context, demonstrate that our approach surpasses state-of-the-art task-conditional methods.
\end{abstract}
\section{Introduction}
Dense prediction tasks, such as semantic segmentation and depth estimation, predict pixel-wise labels and play a crucial role in computer vision research. Recently, there has been a growing interest in learning multiple dense prediction tasks simultaneously. To tackle this multi-task dense prediction problem, the Multi-Task Learning (MTL) framework has gained prominence. 
MTL trains a single model to learn shared representations across tasks, effectively capturing common and complementary information, thereby enhancing overall performance \cite{survey, aich2023efficient}. Moreover, MTL's parameter sharing mechanism increases model efficiency compared to the single-task scenario, where separate models are dedicated to each task.

Encoder-focused \cite{cross-stitch, nddr-cnn, mtan} and decoder-focused methods \cite{pad-net, mti-net, atrc, invpt, shala} are two primary classes for multi-task frameworks \cite{survey}, and both of them design individual branches for each task. Meanwhile, an alternative direction in MTL is task-conditional model \cite{astmt, rcm, tsn, composite}. This approach comprises a shared stem and task-specific modules instead of multiple specialized branches. Unlike conventional methods that generate results for several tasks at one time, task-conditional methods perform forwarding separately by activating the corresponding modules for each task. Consequently, the features adapt from the shared modules to suit the specific domain of each task. Our proposed method follows this approach to achieve advantages in parameter efficiency and architectural flexibility, making it more practical for real-world applications \cite{rcm}.

Existing task-conditional methods have a common shortage in that their performance is constrained because they are based on Convolutional Neural Networks (CNN), which model spatial and task-related features within relatively localized receptive fields \cite{nonlocal}. Vision Transformer (ViT) \cite{transformer, vit} models have been introduced to address dense prediction problems in both single-task \cite{swin, pvt, dpt} and multi-task learning \cite{invpt, mtformer, mqt, demt, vpt} settings. Their capability of capturing long-range dependencies has been proved critical for pixel-wise MTL \cite{invpt, mtformer}. Nevertheless, integrating transformers into task-conditional models while maintaining computation efficiency poses a challenge, as ViT-based models considerably add parameters \cite{invpt}. 
Another limitation is the lack of multi-scale feature interaction in the decoding stage. They either utilize simple upsampling and concatenation for each task \cite{astmt, rcm} or progressively upsample and refine the features in a UNet-like manner \cite{tsn, composite}. The former approaches lack adaptive fusion and interaction of feature maps across different scales and fail to learn shared and task-specific information jointly. Conversely, the latter methods involve feature interaction only between adjacent scales and introduce significant computational complexity due to increasing resolution and high dimension.

To tackle the aforementioned limitations, we propose a novel method called Task Indicating Transformer (TIT). TIT aims to address the key challenge of capturing shared representations and task-specific features with long-range dependencies modeling, while preserving parameter efficiency within task-conditional model. 
Specifically, we introduce the Mix Task Adapter module, inspired by the Adapter module's success in transfer learning for transformers \cite{adapter, hypernetwork, polyhistor, vitadapter}. 
Instead of allocating a separate Adapter module for each task, we designate a Task Indicating Matrix while sharing most of parameters across all tasks. This allows the module to learn task-specific representations explicitly using the Task Indicating Matrix, while modeling task-invariant information and cross-task interactions implicitly through the shared components.
This design also emphasizes parameter efficiency as the two heavy projection matrices are decomposed into two pairs of lower-rank matrices. Thus, adapting the model to a new task merely requires replacing a tiny matrix. Furthermore, we propose the Task Gate Decoder module to enhance multi-scale feature interaction guided by task information. We introduce a learnable Task Indicating Vector for generating dense task embeddings and employ the gating mechanism \cite{gru} to learn a reset gate and an update gate. These gates adaptively integrate task embeddings with the fused feature map from the encoder, leading to improved multi-scale feature interaction and refinement.

\begin{figure*}[t]
    \centering
    \setlength{\abovecaptionskip}{1pt}
    \includegraphics[width=0.9\linewidth]{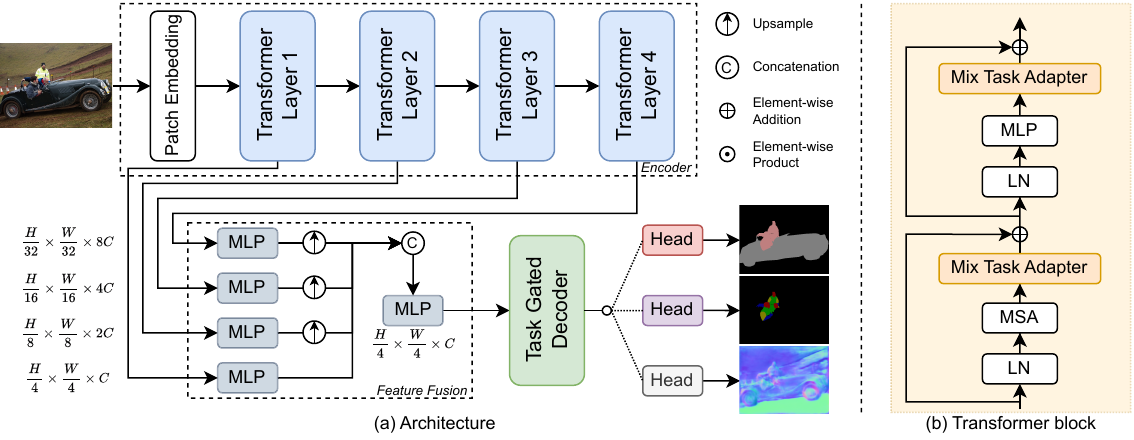}
    \caption{(a) Architecture of proposed Task Indicating Transformer (TIT). (b) Structure of transformer block within the encoder's transformer layers. Mix Task Adapter modules are inserted after the Multi-head Self Attention (MSA) layer and the Multi-Layer Perceptron (MLP) layer.}
    \label{fig:arch}
    \vspace{-0.6cm}
\end{figure*}

In summary, the main contributions of this paper are as follows: 
\begin{itemize}[topsep=0pt,leftmargin=15pt,parsep=0pt,itemsep=2pt]
\item We propose Task Indicating Transformer (TIT), a lightweight task-conditional framework that leverages transformers to capture long-range dependencies and employs the efficient Mix Task Adapter module for feature adaptation and joint learning of intra- and inter-task information via Task Indicating Matrix.
\item We introduce the Task Gate Decoder module, which enables multi-scale feature interaction and refinement conditioned by tasks. The module learns Task Indicating Vectors and controls the adaptive integration of task embeddings and the fused feature map using the gating mechanism.
\item Experiments on two widely used benchmarks, NYUD-v2 and PASCAL-Context, demonstrate that the proposed model outperforms previous state-of-the-art methods in task-conditional dense predictions.
\end{itemize}

\section{Methodology}


\subsection{Preliminary}
Let $\x\in[0,255]^{H\times W\times 3}$ represents an input RGB image, where $H$ and $W$ denote the image's height and width, respectively. For $N$ dense prediction tasks $T=\{t_1, t_2, \dots, t_N\}$, $\y_t\in \mathbb{R}^{H\times W \times O_t} (t\in T)$ stands for the expected prediction for task $t$, where $O_t$ indicates the number of output channels specified by the task. For example, $O_t=3$ in surface normal estimation, while in edge detection, $O_t=1$.
For conventional MTL methods, we employ a neural network $f_\phi$, which outputs predictions for all $N$ tasks concurrently:
\begin{equation}
\setlength{\abovedisplayskip}{3pt}
    \setlength{\belowdisplayskip}{3pt}
    f_\phi(\x)=\bigcup_{t\in T} \y_t.
\end{equation}
Alternatively, in a task-conditional model, each task is performed through a separate forward pass, formulated as:
\begin{equation}
\setlength{\abovedisplayskip}{3pt}
    \setlength{\belowdisplayskip}{3pt}
    f_\phi(\x, t)=\y_t, \ \forall t\in T.
\end{equation}

\subsection{Model Architecture}
As illustrated in Fig. \ref{fig:arch}(a), our model follows an encoder-decoder architecture. The encoder harnesses a vision transformer to extract multi-scale features from four hierarchical layers. To enhance the modeling of long-range dependencies, we employ the well-designed Swin Transformer \cite{swin} as the encoder's backbone. Initially, the input image is partitioned into non-overlapping $4\times 4$ patches, which are then projected into tokens of dimension $C$ within the patch embedding layer. Across each transformer layer, the token count is reduced by 1/4, while the dimension is doubled through patch merging operations. Consequently, the produced representations span from $\frac{H}{4}\times \frac{W}{4}\times C$ to $\frac{H}{32}\times \frac{W}{32}\times 8C$, a configuration proved to be better suited for dense vision tasks \cite{swin, pvt, dpt}. The transformer layers are constructed utilizing the transformer block depicted in Fig. \ref{fig:arch}(b), which additionally includes two Mix Task Adapter modules, serving to facilitate task-conditional learning.

\begin{figure}[t]
    \setlength{\abovecaptionskip}{1pt}
    \centering
    \includegraphics[width=0.9\linewidth]{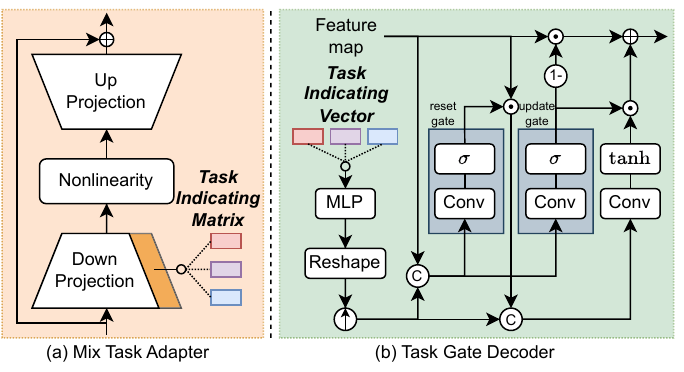}
    \caption{Illustrations of proposed Mix Task Adapter module and Task Gate Decoder module. Different colors of the Task Indicating Matrix and Task Indicating Vector correspond to distinct task types.}
    \label{fig:module}
    \vspace{-0.6cm}
\end{figure}

\begin{table*}[t]
    \setlength{\tabcolsep}{6pt}
    \setlength{\abovecaptionskip}{0pt}
    \centering
    \caption{Comparison with state-of-the-art models on NYUD-v2 dataset. '$\uparrow$' means higher is better and '$\downarrow$' means lower is better. \textbf{Bold text} highlights the best result, while \underline{underlined text} represents the second best result. '\textcolor{red}{$\uparrow$}' denotes our improvement over the second best result.}
    \begin{tabular}{cccccccccc}
    \hline
        Model & Backbone & Params (M) & SemSeg (mIoU)$\uparrow$ & Depth (RMSE)$\downarrow$ & Normals (mErr)$\downarrow$ & Edge (odsF)$\uparrow$ & $\Delta_m\%\uparrow$ \\
    \hline
        Single-task & Swin-T & 110.9 & 41.94 & 0.6112 & 20.11 & 77.33 & 0.00 \\
        Multi-decoder & Swin-T & 28.3 & 40.98 & 0.6283 & 20.22 & 77.02 & -1.51 \\
    \hdashline
        ASTMT \cite{astmt} & ResNet-50 & 45.0 & 32.16 & \underline{0.5700} & 23.18 & 74.50 & -8.88 \\
        RCM \cite{rcm} & ResNet-18 & 39.0 & 34.20 & \underline{0.5700} & 22.41 & 68.44 & -8.66 \\
        RCM \cite{rcm} & ResNet-34 & 53.7 & \underline{36.19} & \textbf{0.5500} & \underline{21.70} & 69.50 & \underline{-5.43} \\
        TSN \cite{tsn} & ResNet-18 & 18.3 & 25.90 & 0.7270 & 26.10 & 67.90 & -24.79 \\
        TSN \cite{tsn} & Swin-T & 39.2 & 32.38 & 0.6874 & 22.25 & \underline{75.69} & -12.01 \\
        TIT (Ours) & Swin-T & 30.9 & \textbf{41.36} & 0.5925 & \textbf{19.68} & \textbf{77.30} & \textbf{0.94}(\textcolor{red}{$\uparrow 6.37$}) \\
    \hline
    \end{tabular}
    \label{tab:ny}
    \vspace{-0.6cm}
\end{table*}

The feature fusion incorporates dimension reduction adaptively, as the features from the encoder are projected onto the dimension of $C$ via MLP modules. Subsequently, high-level features from layer 2 to layer 4 are upsampled to align with the spatial resolution $\frac{H}{4}\times \frac{W}{4}$ of layer 1. The four feature maps are then concatenated and passed through another MLP layer to reduce the dimension back to $C$. Following this procedure, the feature map undergoes further multi-scale interaction and refinement by the Task Gate Decoder module for improved decoding. Finally, task-specific prediction heads are adopted to yield outputs for each task. Each head comprises two upsampling layers and an $1\times 1$ convolutional layer for channel projection.
\vspace{-0.2cm}

\subsection{Mix Task Adapter}
The Adapter \cite{adapter} is a bottleneck-like module placed within transformer block, enabling parameter-efficient fine-tuning and adaptation. The Adapter consists of three main components: a down projection layer $\W_\text{down}\in \mathbb{R}^{d\times n}$, a nonlinearity function $\delta(\cdot)$, and an up projection layer $\W_\text{up}\in \mathbb{R}^{n\times d}$, where $n=\alpha d$ and $\alpha < 1$ represents a constant signifying the projection ratio. A residual connection from the input $\x_\text{in}\in \mathbb{R}^d$ is used to obtain the output $\x_\text{out}\in \mathbb{R}^d$, which can be formulated as:
\begin{equation}
    \setlength{\abovedisplayskip}{3pt}
    \setlength{\belowdisplayskip}{3pt}
    \x_\text{out}=\delta(\x_\text{in}\W_\text{down})\W_\text{up}+\x_\text{in}.
\end{equation}

Approaches based on Adapter have shown promising results in multi-task adaptation for both NLP and vision tasks \cite{adapter, hypernetwork, polyhistor, vitadapter}, and previous task-conditional method has also used a bypass structure called 'residual adapter' \cite{astmt}. Nevertheless, these works typically use a set of task-specific Adapter modules for each task and switch among different sets to perform distinct tasks.
As Adapters are trained independently, this paradigm fails to exploit shared and cross-task information among multiple tasks. Additionally, it inadvertently results in a substantial increase in model parameters, which becomes particularly problematic when dealing with feature vectors of high dimensionality (\textit{e.g.}, 768 in the fourth layer of the Swin-T model) and when processing a considerable number of tasks.

To address these limitations, we design a Task Indicating Matrix in our Mix Task Adapter module, as depicted in Fig. \ref{fig:module}(a), which can introduce task-related guidance into the Adapter. We decompose the down projection matrix and up projection matrix into two pairs of lower-rank matrices $\W^p$ and $\W^q$ as follows:
\vspace{-0.1cm}
\begin{gather}
\setlength{\abovedisplayskip}{0pt}
\setlength{\belowdisplayskip}{0pt}
    \W_\text{down}=\W_\text{down}^p \W_\text{down}^q, \\
    \W_\text{down}^p\in\mathbb{R}^{ d\times m}, \W_\text{down}^q\in \mathbb{R}^{m\times n}, \\
    \W_\text{up}=\W_\text{up}^q \W_\text{up}^p, \\
    \W_\text{up}^p\in\mathbb{R}^{m \times d}, \W_\text{up}^q\in \mathbb{R}^{n \times m},
\end{gather}
where $m<n$ is another hyper-parameter that can be adjusted. For effective task-conditional learning, we exclusively designate $\W_\text{down}^q$ as the task-specific Task Indicating Matrix, while the remaining three matrices are shared among tasks. Thus, task-specific and task-agnostic components are \textit{mixed} in the module, where targeted representations are learned by inserting different Task Indicating Matrix into the module, while universal characteristics and cross-task correlations are modeled by the shared matrices.

In this way, the dilemma in parameter usage can be optimized since the number of parameters in a single Adapter module can be reduced from $2nd^2$ to $2m(n+d)$. As an illustrative example, assuming $\alpha=1/4$ and $m=n/2$, this reduction amounts to a 37.5\% decrease in parameters. Moreover, when considering $N$ tasks, the parameter count using $N$ separate Adapter modules totals $2Nnd^2$, whereas our Mix Task Adapter module requires only $(N+1)mn+2md$ parameters, further underscoring its efficiency in parameter utilization.

\subsection{Task Gate Decoder}
In order to enhance multi-scale feature interaction and task-related refinement in the decoder, we draw inspiration from GRU (Gated Recurrent Unit) \cite{gru} and ConvGRU \cite{convgru}, and design the Task Gate Decoder module, as shown in Fig. \ref{fig:module}(b). GRU and ConvGRU were originally developed for Recurrent Neural Networks (RNN) to process sequence data such as times series or natural language sentences. They maintain a hidden state that captures information about previous elements and update it with new inputs. We leverage their ability to update the fused feature map, which serves as the hidden state in our module. The gating mechanism controls the flow of information in and out of the feature map and enables adaptive information update and forgetting. Additionally, we introduce a trainable Task Indicating Vector to provide task-specific guidance by integrating task embeddings as input to the gates.

The Task Indicating Vector $\v^t\in \mathbb{R}^{k\times 1}$ of task $t$ is first passed through a shared MLP layer, reshaped, and upsampled 8 times to match the resolution of the fused feature map, which is $\frac{H}{4}\times \frac{W}{4}$:
\vspace{-0.2cm}
\begin{align}
\setlength{\abovedisplayskip}{0pt}
\setlength{\belowdisplayskip}{0pt}
    \hat{\v}^t&=\text{MLP}(\v^t)\in  \mathbb{R}^{(H/32)\cdot (W/32) \cdot C_T\times 1}, \\
    \hat{\e}^t&=\text{Reshape}(\hat{\v}^t) \in \mathbb{R}^{\frac{H}{32}\times \frac{W}{32}\times C_T}, \\
    \e^t &=\text{Upsample}(\hat{\e}^t)\in \mathbb{R}^{\frac{H}{4}\times \frac{W}{4}\times C_T},
\end{align}
where $C_T$ is the dimension of the generated task embedding $\e^t$.

Similar to GRU, our module has two gates: a reset gate $\r$ determines how much of the input feature should be forgotten and how much of the task embedding should be considered when updating the feature, and an update gate $\z$ controls how much of the input feature should be carried over to the task-specific output. These two gates are computed by two convolutional layers:
\vspace{-0.2cm}
\begin{align}
\setlength{\abovedisplayskip}{0pt}
\setlength{\belowdisplayskip}{0pt}
    \r^t&=\sigma(\text{Conv}_\text{r}([\e^t, \x^t])), \\
    \z^t&=\sigma(\text{Conv}_\text{z}([\e^t, \x^t])),
\end{align}
where $\sigma$ is the logistic sigmoid function and $[\cdot,\cdot]$ is concatenation. 
The final output of the Task Gate Decoder is then computed by
\vspace{-0.1cm}
\begin{align}
\setlength{\abovedisplayskip}{0pt}
\setlength{\belowdisplayskip}{0pt}
    \tilde{\x}^t&=\tanh((\text{Conv}_\text{o}([\e^t, \r^t\odot \x^t]))), \\
    \hat{\x}^t&=(1-\z^t)\odot \x^t+\z^t\odot \tilde{\x}^t,
\end{align}
where $\text{Conv}_\text{o}$ is the output convolutional layer and $\odot$ denotes element-wise product. Notably, all convolutional layers are shared among tasks, capitalizing on the parameter sharing mechanism.

\begin{table*}[t]
    \setlength{\tabcolsep}{3pt}
    \setlength{\abovecaptionskip}{0pt}
    \centering
    \caption{Comparison with state-of-the-art models on PASCAL-Context dataset.}
    \begin{tabular}{ccccccccccc}
    \hline
        Model & Backbone & Params (M) & SemSeg (mIoU)$\uparrow$ & Parts (mIoU)$\uparrow$ & Sal (mIoU)$\uparrow$ & Normals (mErr)$\downarrow$ & Edge (odsF)$\uparrow$ & $\Delta_m\%\uparrow$\\
    \hline
        Single-task & Swin-T & 138.6 & 70.47 & 66.21 & 64.82 & 13.45 & 75.78 & 0.00 \\
        Multi-decoder & Swin-T & 28.5 & 64.65 & 59.77 & 64.45 & 13.95 & 72.85 & -5.23 \\
    \hdashline
        ASTMT \cite{astmt} & ResNet-26 & 31.3 & 64.61 & 57.25 & 64.70 & 15.00 & 71.00 & -7.97\\
        ASTMT \cite{astmt} & ResNet-50 & 49.4 & \underline{68.00} & \underline{61.12} & 65.71 & 14.68 & 72.40 & \underline{-4.68} \\
        RCM \cite{rcm} & ResNet-18 & 46.1 & 65.70 & 58.12 & \textbf{66.38} & \textbf{13.70} & 71.34 & -4.86\\
        TSN \cite{tsn} & ResNet-34 & 28.4 & 67.60 & 58.00 & 64.30 & 16.10 & 71.80 & -8.45\\
        TSN \cite{tsn} & Swin-T & 39.1 & 67.30 & 61.11 & 64.29 & 14.55 & \textbf{74.04} & -4.70\\
        TIT (Ours) & Swin-T & 31.3 & \textbf{70.04} & \textbf{62.68} & \underline{66.14} & \underline{14.43} & \underline{73.91} & \textbf{-2.73}(\textcolor{red}{$\uparrow 1.95$})\\
    \hline
    \end{tabular}
    \label{tab:pc}
    \vspace{-0.6cm}
\end{table*}

\begin{table}[t]
    \setlength{\tabcolsep}{5pt}
    \setlength{\abovecaptionskip}{0pt}
    \centering
    \caption{Effectiveness of different components in proposed approach. `ST' stands for single-task baseline. `MTA' denotes Mix Task Adapter and `TGD' denotes Task Gate Docoder.}
    \begin{tabular}{ccccccc}
    \hline
        Model & SemSeg$\uparrow$ & Depth$\downarrow$ & Normals$\downarrow$ & Edge$\uparrow$ & $\Delta_m\%\uparrow$\\
    \hline
        ST & 41.94 & 0.6112 & 20.11 & 77.33 & 0.00 \\
    \hdashline
        +MTA & 39.61 & 0.6644 & 20.78 & 76.36 & -4.71 \\
        +TGD & 39.32 & 0.6420 & 19.68 & \textbf{77.36} & -2.28 \\
        TIT full & \textbf{41.36} & \textbf{0.5925} & \textbf{19.68} & 77.30 & \textbf{0.94} \\
    \hline
    \end{tabular}
    \label{tab:ab}
    \vspace{-0.6cm}
\end{table}

\section{Experiments}

\subsection{Experimental Setup}
\textbf{Datasets}
We use two widely used benchmark datasets for multi-task dense predictions: NYUD-v2 \cite{nyud} and PASCAL-Context \cite{pc}. The NYUD-v2 dataset consists of 795 training and 654 testing images of indoor scenes, with annotations for four tasks: semantic segmentation ('SemSeg'), depth estimation ('Depth'), surface normal estimation ('Normals'), and edge detection ('Edge'). The PASCAL-Context dataset contains 4998 training and 5105 testing images and includes annotations for five tasks: semantic segmentation, human parts segmentation ('Parts'), saliency estimation ('Sal'), surface normal estimation and edge detection. We follow a typical data augmentation pipeline, including scaling, cropping, horizontal flipping, and color jittering, consistent with existing methods \cite{rcm, tsn}.

\noindent\textbf{Implementation}
We utilize a Swin Transformer \cite{swin} backbone (Swin-T) pre-trained on ImageNet-22K. In the Mix Task Adapter modules, we set $\alpha=1/4$ and $m=n/2$, while in the Task Gate Decoder module, we use $k=64$ and $C_T=16$.
The model is trained for 500 epochs on NYUD-v2 and 300 epochs on PASCAL-Context, with a batch size of 8 and 32, respectively. We use the AdamW optimizer with a learning rate of 1e-4 and a weight decay rate of 1e-4. All experiments are conducted on 4 NVIDIA RTX3090 GPUs.

\noindent\textbf{Baselines}
We construct strong single-task baselines and multi-decoder baselines. These baselines share the same architecture and backbone as our TIT, except that they do not use our proposed modules. The single-task baselines train individual networks for each task, while the multi-decoder baselines use task-specific feature fusions and prediction heads.

\noindent\textbf{Evaluation metric}
We follow the standard practice in evaluation metrics \cite{astmt, rcm, tsn}. We use mean Intersection over Union (mIoU) for semantic segmentation, human parts segmentation, and saliency detection, mean angular error (mErr) for surface normal estimation, Root Mean Square Error (RMSE) for depth estimation, and optimal-dataset-scale F-measure (odsF) for edge detection.
To quantify overall multi-task performance, we calculate the average per-task performance drop ($\Delta_m$) with respect to the single-task baseline, as established in prior work \cite{tsn}. 
\vspace{-0.2cm}

\subsection{Experimental Results}
\textbf{Comparison to state-of-the-art}
Table \ref{tab:ny} and Table \ref{tab:pc} present the performance of our method in comparison to existing task-conditional approaches, namely ASTMT \cite{astmt}, RCM \cite{rcm}, and TSN \cite{tsn}. To ensure a fair comparison, we select models with backbones that have a similar or greater number of parameters than ours. Moreover, we re-implement TSN using the Swin-T backbone and report its performance. Overall, our method consistently outperforms the multi-decoder baseline and existing approaches by a significant gap, showcasing the best average performance. 
Notably, our method exhibits superior performance compared to TSN with the same Swin-T backbone on NYUD-v2, with an improvement of around 13\%. These results clearly prove the superior capability of task-conditional representation learning attained by our well-designed modules.

\begin{table}[t]
    \setlength{\tabcolsep}{2.5pt}
    \setlength{\abovecaptionskip}{0pt}
    \centering
    \caption{Impact of matrix dimension $m$ in the Mix Task Adapter module, compared to applying original Adapter modules for each task. 'Params' means the number of parameters in all Adapter or Mix Task Adapter modules.}
    \begin{tabular}{ccccccc}
    \hline
        Module & Params & SemSeg$\uparrow$ & Depth$\downarrow$ & Normals$\downarrow$ & Edge$\uparrow$ & $\Delta_m\%\uparrow$\\
    \hline
        ST & - & 41.94 & 0.6112 & 20.11 & 77.33 & 0.00 \\
    \hdashline
        Adapter & 8.68M & \textbf{41.59} & \textbf{0.5797} & 20.01 & 77.20 & \textbf{1.16} \\
        $m=n/8$ & \textbf{0.46}M & 40.71 & 0.5934 & \underline{19.79} & \underline{77.23} & 0.36 \\
        $m=n/4$ & 0.90M & 40.85 & \underline{0.5912} & 19.85 & 77.09 & 0.41 \\
        $m=n/2$ & 1.78M & \underline{41.36} & 0.5925 & \textbf{19.68} & \textbf{77.30} & \underline{0.94} \\
    \hline
    \end{tabular}
    \label{tab:ab2}
    \vspace{-0.35cm}
\end{table}

\begin{figure}[t]
    \setlength{\abovecaptionskip}{3pt}
\scriptsize{\hspace{6.5mm}Image\hspace{7mm}SemSeg\hspace{7.5mm}Parts\hspace{10mm}Sal\hspace{8mm}Normals\hspace{7mm}Edge}
    \vspace{-3.5mm}
    
    \subfloat{
        \rotatebox{90}{\scriptsize{~~~~GT}}
        \includegraphics[width=0.155\linewidth]{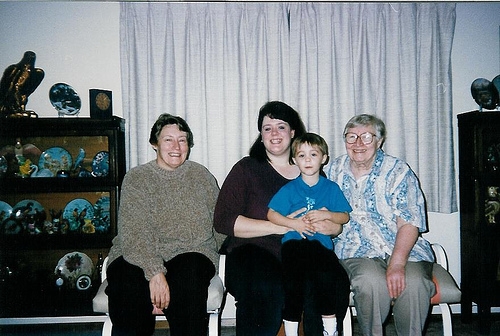}
        \includegraphics[width=0.155\linewidth]{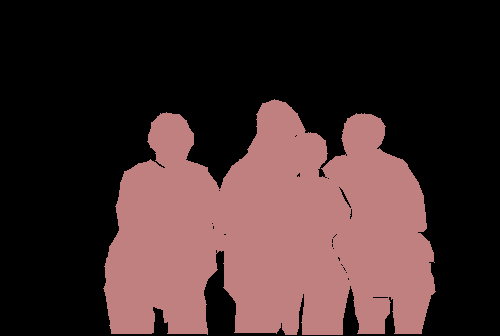}
        \includegraphics[width=0.155\linewidth]{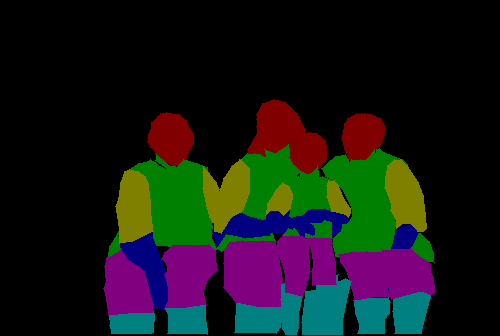}
        \includegraphics[width=0.155\linewidth]{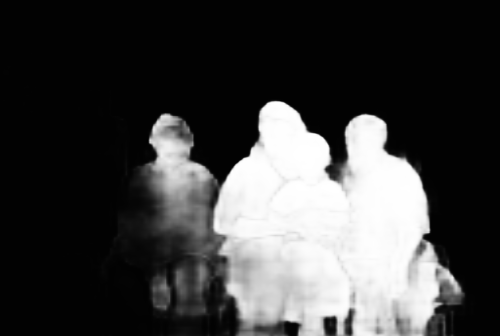}
        \includegraphics[width=0.155\linewidth]{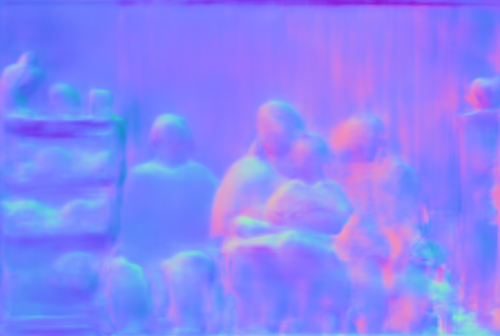}
        \includegraphics[width=0.155\linewidth]{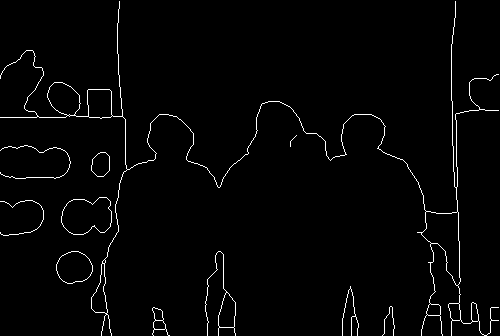}
    }
    \vspace{-3.5mm}
    
    \subfloat{
        \rotatebox{90}{\scriptsize{~~~~TSN}}
        \includegraphics[width=0.155\linewidth]{1034/2008_001034.jpg}
        \includegraphics[width=0.155\linewidth]{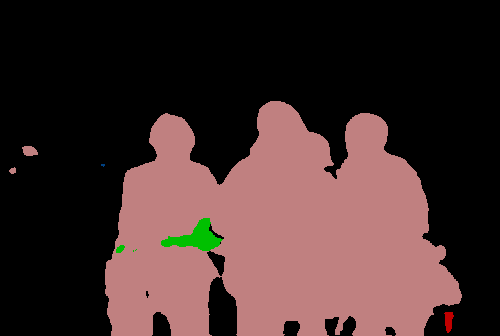}
        \includegraphics[width=0.155\linewidth]{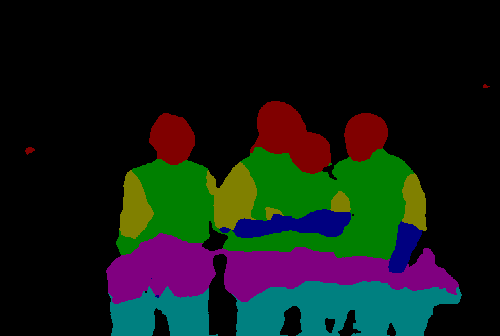}
        \includegraphics[width=0.155\linewidth]{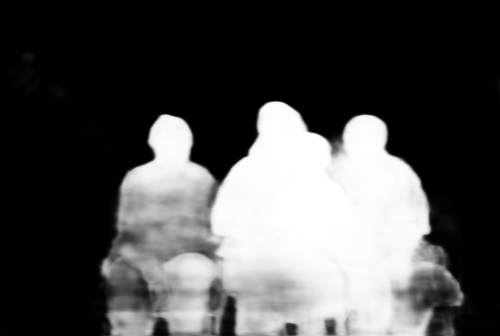}
        \includegraphics[width=0.155\linewidth]{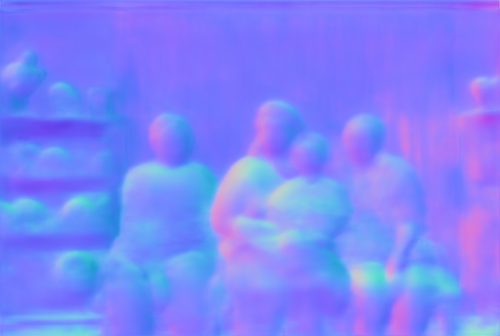}
        \includegraphics[width=0.155\linewidth]{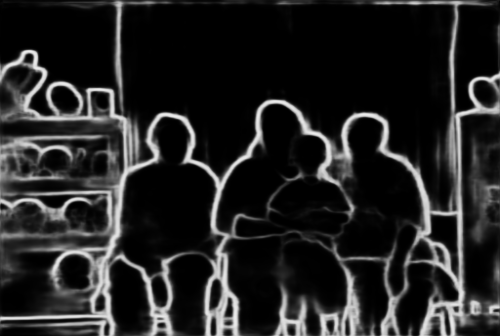}
    }
    \vspace{-3.5mm}

    \subfloat{
        \rotatebox{90}{\scriptsize{~~~~TIT}}
        \includegraphics[width=0.155\linewidth]{1034/2008_001034.jpg}
        \includegraphics[width=0.155\linewidth]{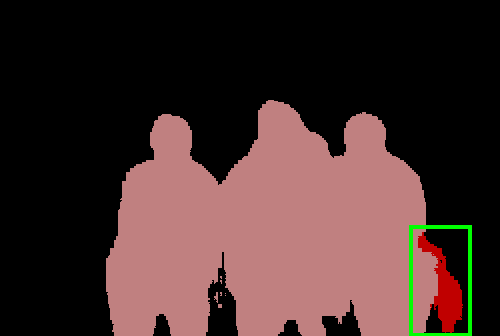}
        \includegraphics[width=0.155\linewidth]{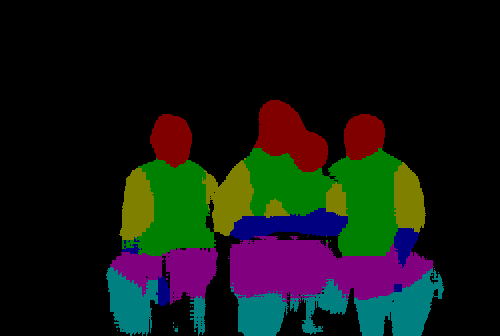}
        \includegraphics[width=0.155\linewidth]{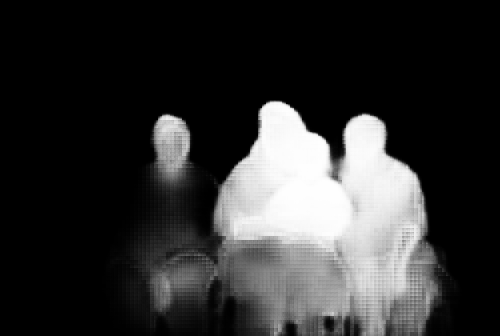}
        \includegraphics[width=0.155\linewidth]{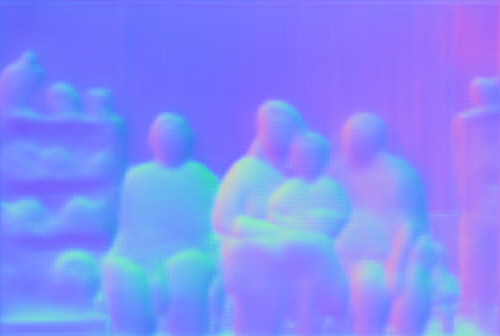}
        \includegraphics[width=0.155\linewidth]{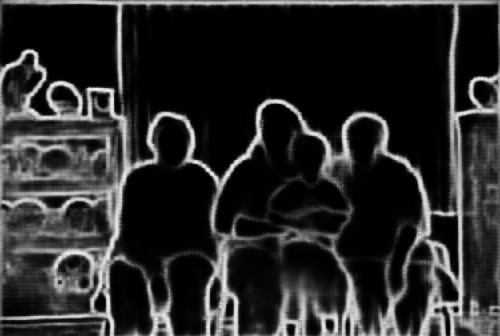}
    }
    \caption{Qualitative results on PASCAL-Context dataset.}
    \label{fig:qual}
    \vspace{-0.7cm}
\end{figure}

\noindent\textbf{Ablation study} We conduct ablation studies to validate the effectiveness of our proposed modules on NYUD-v2. Table \ref{tab:ab} clearly demonstrates that both the Mix Task Adapter and Task Gate Decoder modules contribute to improvements when comparing the full TIT model to using either of them separately. 
Specifically, the Task Gate Decoder results in an overall gain of 5.65\% and the Mix Task Adapter also enhances the average performance by 3.22\%. 
Furthermore, we conduct an analysis to assess how the Mix Task Adapter module performs with varying lower-rank matrix dimensions $m$. We also include a model that applies the original Adapter module for each task under the same setting for comparison. As presented in Table \ref{tab:ab2}, when larger matrix sizes are used, the improvements over single-task setting become more pronounced. It is worth noting that our module achieves comparable performance with the Adapter while utilizing only 20\% of the parameters. It can even improve parameter efficiency by 95\%, albeit at a slight cost of a 0.8\% performance drop.

\noindent\textbf{Qualitative results} In Fig. \ref{fig:qual}, we show a qualitative comparison between the predictions generated by our model, TSN, and the ground truth. Intuitively, our model produces results that more closely resemble the ground truths than competitor. Additionally, we successfully segment the chair, which is not manually annotated in the label, as indicated by the green bounding box.
\vspace{-0.2cm}

\section{Conclusion}
This paper presents a novel architecture termed Task Indicating Transformer for task-conditional dense predictions. We incorporate a Mix Task Adapter module within the transformer structure to enhance global dependency modeling and parameter-efficient feature adaptation. We also propose a Task Gate Decoder module, which enables task-guided adaptive multi-scale feature interaction and refinement. Through extensive experiments conducted on the NYUD-v2 and PASCAL-Context benchmarks, we substantiate the effectiveness of our method, underscoring its superiority over state-of-the-art methods in this field. Our future research will focus on dynamic balancing of task losses and gradients under task-conditional paradigm and continual enhancement of model efficiency and applicability.

\bibliographystyle{IEEEbib}
\bibliography{egbib}

\end{document}